# Examining the Capability of GANs to Replace Real Biomedical Images in Classification Models Training

Vassili Kovalev and Siarhei Kazlouski

United Institute of Informatics Problems, Surganova St, 6, 220012 Minsk, Belarus
{vassili.kovalev, kozlovski.serge}@gmail.com

**Abstract.** In this paper, we explore the possibility of generating artificial biomedical images that can be used as a substitute for real image datasets in applied machine learning tasks. We are focusing on generation of realistic chest X-ray images as well as on the lymph node histology images using the two recent GAN architectures including DCGAN and PGGAN. The possibility of the use of artificial images instead of real ones for training machine learning models was examined by benchmark classification tasks being solved using conventional and deep learning methods. In particular, a comparison was made by replacing real images with synthetic ones at the model training stage and comparing the prediction results with the ones obtained while training on the real image data. It was found that the drop of classification accuracy caused by such training data substitution ranged between 2.2% and 3.5% for deep learning models and between 5.5% and 13.25% for conventional methods such as LBP + Random Forests.

## 1       Introduction

Machine learning (ML) and in particular deep learning (DL) techniques become more and more popular in biomedical image analysis domain during the last several years [1]. Despite of a high efficiency of these approaches (see, for example,[1, 2]) the success of ML methods strongly depends on the availability of large training datasets accompanied by appropriate annotations. At the same time, data specificity clearly implies a number of ethical, legal, and technical challenges in retrieving and sharing datasets of biomedical images. The frequency of natural appearance of patients with particular pathologies limits the possible number of cases with certain pathologies. The image data collecting procedure may be time consuming and often requires advanced equipment. On the top, the data annotation supposes involvement of experienced specialists. This could be costly or even not feasible at all due to a very high cost. Also, ethical and privacy regulations make it difficult or, in some particular cases, entirely impossible to distribute collected biomedical image data publicly as well as between some institutions, or even on-site [3, 4].

Although a number of projects (e.g., National Biomedical Imaging Archive**,** OASIS Brains project, OpenNeuro.org and some other) tend to provide public image datasets of some modalities, the problem of data scarcity remains and it constrains possibilities of researchers in the field of biomedical imaging.

In this work, we study the possibility to overcome the problem of biomedical data scarcity by applying modern SOTA image synthesis technique – Generative Adversarial Networks (GANs) [5] for creation of artificial biomedical image datasets that can be used as a publicly available substitute for real images in applied ML tasks.

Recent researches that used GANs in biomedical domain have proven the efficiency of this approach in modality-to-modality translation, image de-noising, reconstruction, enhancement, segmentation and image synthesis itself [6, 7]. Existing works related to the problem of image synthesis can be split into two groups. A majority of works belongs to the first group which puts emphasis on utilizing GANs for augmentation of some existing image dataset as well as transformation and widening them in order to achieve more accurate results in various applied task. The works of the second group aim to replace real image data with artificial ones.

This study belongs to the second group and we are trying to widen the existing research works in the field by involving multiple combinations of data types and methods. In particular, we are working with the two large image datasets of opposite nature, namely, the chest X-rays and histology images. For artificial image synthesis, we are using the following two GAN architectures: the Deep Convolutional generative adversarial networks [8] (DCGAN) and the Progressive Growing generative adversarial networks [9] (PGGAN). Also, we utilizing two different approaches for the quantitative evaluation of the quality of artificial images which include training of convolutional networks for solving image classification tasks and using conventional ML models based on feature extraction followed by traditional classifiers.

Hence, with this work we are trying to provide the results of the assessment of the quality of artificial images as well as to enrich the paper content by way of reporting interesting observations regarding the two GAN architectures mentioned above.

## 2     Original Image Data

Two kinds of biomedical images were selected in order to cover images of two different modalities of opposite nature. The first image type is represented by chest X-ray images which are grayscale and are holding certain anatomical shape with the relatively high role of spatial, "geometrical" structure. The second image type is represented by histological images which are, on the contrary, color and can be viewed as a "shape-free" random color texture. Both image types are important for medical practice: x-ray images are often used in screening for detecting lung, skeletal and cardiovascular system abnormalities as well as for monitoring various treatment processes; histological images are continuously playing the role of a gold standard in cancer diagnosis.

### 2.1 Chest X-ray Images

X-ray image data used in this study were the natively-digital X-ray scans extracted from a PACS system, containing results of a periodic chest screening of adult population of a two-million city. The version of the database we used here contains a total of 1,908,926 X-ray images accompanied by data records which include patients' ID, age, gender and textual radiological reports made by a chief radiologist.

Technically, all the X-ray scans were represented by a one-channel 16-bit non-compressed images. The original image resolution varied from a rarely occurred 520×576 pixels to the size of 2800×2531 pixels. In order to unify dataset and make it usable in our research, images were resized to 512 x 512 pixels and normalized by the lungs convex hull using 5 and 95 percentiles of brightness. Images were labeled with patient age, gender and abnormality (if any) what was extracted from radiological reports.

**Normal Chest X-Ray (X-Norm)** included images without visible signs of any type of abnormalities in mediastinum, skeleton and the lungs themselves. A total of 566,712 chest images were included in this image dataset after a technical clearance.

### 2.2 Histology Images

Histology images used in this study were images of lymph node [10]. The source dataset was composed of 400 whole-slide images of sentinel lymph node, 270 of which were labeled as normal regions and the rest 130 as the regions containing metastases from breast cancer. Technically, all the images were presented by color 3-channel images of tissue samples with resolution up to 200 000 pixels in both width and height. In order to retrieve manageable-sized data, original images were cut into small 256 x 256 regions, which are referred to as the image "tiles". Each tile inherited a label from its source whole-slide image. The following two data sets were extracted:

**Normal Histology (H-Norm)** included 50 000 randomly chosen tiles taken from normal regions.

**Tumor Histology (H-Tumor)** included 50 000 randomly chosen tiles taken from regions containing metastases.

### 2.3 Study groups

The image study groups created based on the above mentioned X-ray and histology images which are used in all the experiments are presented in Table 1 along with their major characteristics.

As it can be seen from the Table, the X-Norm-L study group uses not evenly-binned age subgroups. This is caused by a non-equivalent distribution of gender and ages in the original dataset. Due to a well-known social reasons, there is a lack of scans of elderly people. This is especially true for male subjects. In order to maintain the maximum coverage of ages and still get a large, well- balanced dataset, we decided to use non-homogeneous age grouping with smaller group size for younger subjects (what is mostly <50) and larger group size for elderly ages. In order to ensure that this kind of grouping will not cause quality drop in experimental result, a regularly-sampled age groups were also utilized in this study.

Table 1. Study groups and their characteristics

| Group Name | Subset | Size | Comments |
|---|---|---|---|
| X-Norm | Train | 120 570 | Images labeled by gender-age group. Age groups are: 18-19, 20-21, … , 67-68. Balanced by gender and age (30 groups). |
| X-Norm | Test | 13 410 | |
| X-Norm-L | Train | 71 026 | Images labeled by gender-age group. Age groups are: 18-19, 20-21, 22-23, 24-25, 26-27, 28-30, 31-33, 34-36, 37-39, 40-44, 45-49, 50-54, 55-60, 61-66, 67-77. Balanced by gender in 34 age groups. |
| X-Norm-L | Test | 7 922 | |
| H-Norm | Train | 40 000 | None |
| H-Norm | Test | 10 000 | |
| H-Tumor | Train | 40 000 | |
| H-Tumor | Test | 10 000 | |

## 3   Experimental Setup

### 3.1   Image Generation

The open-source implementations of two GAN architectures were used as starting points for generation of image datasets including DCGAN [11] and PGGAN [12]. Each model was applied to the training set of each study group we created.

**X-ray.** Due to the presence of multiple gender-age groups in the dataset and relatively small amount of samples for each group, we tried both conditional and unconditional model training. In the unconditional setup, a subset of data was extracted from training set of the study group for each label and individual model was trained on this subset without any additional input to discriminator or generator networks. In case of conditional setup, the models were trained on the entire training sets of study group and gender-age label of each image was encoded and passed to generator and discriminator networks during the training. Thus, for unconditional setup a total of 64 DCGAN and 64 PGGAN models were trained. Also, two DCGAN models and two PGGAN models were trained for the conditional setup.

Considering the hardware limitations and based on a visual assessment of generated images during the preliminary experiments, the following setup was used:

Latent vector dimensionality: 512; batch size: 32; input and output image size: 256 x 256 pixels; training epochs: 30 for unconditional setup, 60 for conditional setup; optimizer: Adam. Same batch size and epochs number were used for all layer depths in PGGAN.

**Histology.** For histology data we used unconditional model training. That means that the one model was trained on the entire training sets of study groups without any additional input to discriminator or generator networks. Thus, two DCGAN models and two PGGAN models were trained. The following setup was used:

Latent vector dimensionality: 256; batch size: 32; input and output image size: 256 x 256 pixels; training epochs: 10; optimizer: Adam. Same batch size and epochs number were used for all layer depths in PGGAN.

After completing training of GANs, the training datasets of artificial images were generated using best checkpoints for each model. The best checkpoints selection criteria were based on visual examination of random samples generated for each epoch during the training procedure.

Finally, the artificial versions of original study groups from Table 1 were generated what is resulted in 12 datasets of artificial images. Generated datasets were of the same size and contain the same labels distribution as corresponding original datasets.

### 3.2 Artificial Dataset Assessment

Artificial image datasets were first examined visually and then either discarded as obviously unacceptable or passed to further quantitative assessment stage. In case of the use of DCGAN method for generating X-ray images the invalid results were removed with the help of well-known image hash technique (hash size h=16, hash type=average, distance metric = Hamming) which check images for correct global body shape. Quantitative examination was performed by way of comparing the accuracy achieved on benchmark classification tasks by ML models trained either on real image datasets or on artificial image datasets.

General algorithm of benchmarking can be described as follows:
(a) Formulate the benchmark classification task.
(b) Determine he training sets of real and artificial data and test set of real data.
(c) Determine the model architecture and parameters to use for classification.
For each of selected model:
(d) Train model on the training set of real data.
(e) Train model on the training set of artificial data.
(f) Score models from (d) and (e) on the test set of real data and compare results.

The benchmark tasks and data sets descriptions are presented in Table 2.

**Table 2.** Benchmark classification tasks

| Data type | Classification task | Training set size | Test set size |
|---|---|---|---|
| Histology | By class: norm or tumor | 8000 | 2000 |
| X-ray | By gender: female or male | 8000 | 1000 |
| X-ray | By age group: "young" (18-38) or "mature" (48-68) | 8000 | 1000 |

All the training and the test image datasets were balanced by target class labels. In case of X-ray datasets, they were also balanced by subjects' gender-age group.

The following ML models were used for classification:
(1) VGG16 deep convolutional network; (2) K-NN; (3) SVM; (4) Random forest.

For conventional methods, the LBP descriptors of images with radius 2, 3, 4 as well as their combination were used as image features. In each classification task, the procedure of selection of the best parameters was performed using 5-Fold cross-validation. In this experiment, the color histology images were converted to the grayscale versions in order to use LBP method directly.

For VGG16 network, the whole image datasets were used as network input. The training of the model was performed for 20 epochs with the batch size of 32. The best models were selected based on the best validation loss being obtained.

## 4  Results

**X-ray.** Generally, both DCGAN and PGGAN generators were able to produce visually appealing artificial chest X-ray images (see Fig. 1a and Fig. 2a respectively). However, it was noticed that in about of 40-60% of cases all the examined DCGAN models produced X-ray images with heavily distorted body shape (Fig. 1b). Contrary to DCGAN, the PGGAN models performed much better in reconstructing global image structure due to their inherent property of "multi-resolution", i.e., the ability of gradual, steady image refinement (Fig. 2a). In context of neural networks, this can be also explained by a much more robust handling of macro structure in progressively growing layers of PGGAN.

Our results suggest that PGGAN achieved the best performance in case of conditional training on X-Norm-L while DCGAN created images of the best quality in case of unconditional training on either X-Norm-L or X-Norm. It should be noted that in all the experiments the training process was not very stable and generator models often passed through phases of dramatic image quality drop. In case of PGGAN this drop appeared only on layers which generated images of resolution 128 x 128 and higher.

**Histology.** Here DCGANs have demonstrated faster model collapse than in case of X-ray images. Nevertheless, we were able to generate artificial images of good visual quality on early stage of training, just after 3 epochs for normal histology tiles and after 5 epochs in case of 256x256 tiles of tumor histology images (Fig.1b,c).

Contrary to the DCGANs, we completely failed to generate good quality histology images using PGGAN model. This is because they were always demonstrating very fast model collapse and produced kind of "averaged" histology image patterns like the ones shown in Fig.2b. This happened as early as from the first epoch of training. We conclude that in case of texture-like, randomly structured images progressive image size growth leads to a much faster collapse of training since the first layers generate almost solidly filled images and further layers were not able to restore them.

After determining the best models, which were conditional PGGAN for X-ray images trained on X-Norm-L image dataset and DCGAN for histology images trained on Histology, we generated three artificial image datasets as described in Table 2 and used them to train classification models as described in previous section. A summary of these benchmarking results is presented in Table 3. The results demonstrate that in all the cases the classification accuracy of models trained on artificial data is lower than of models trained on real data.

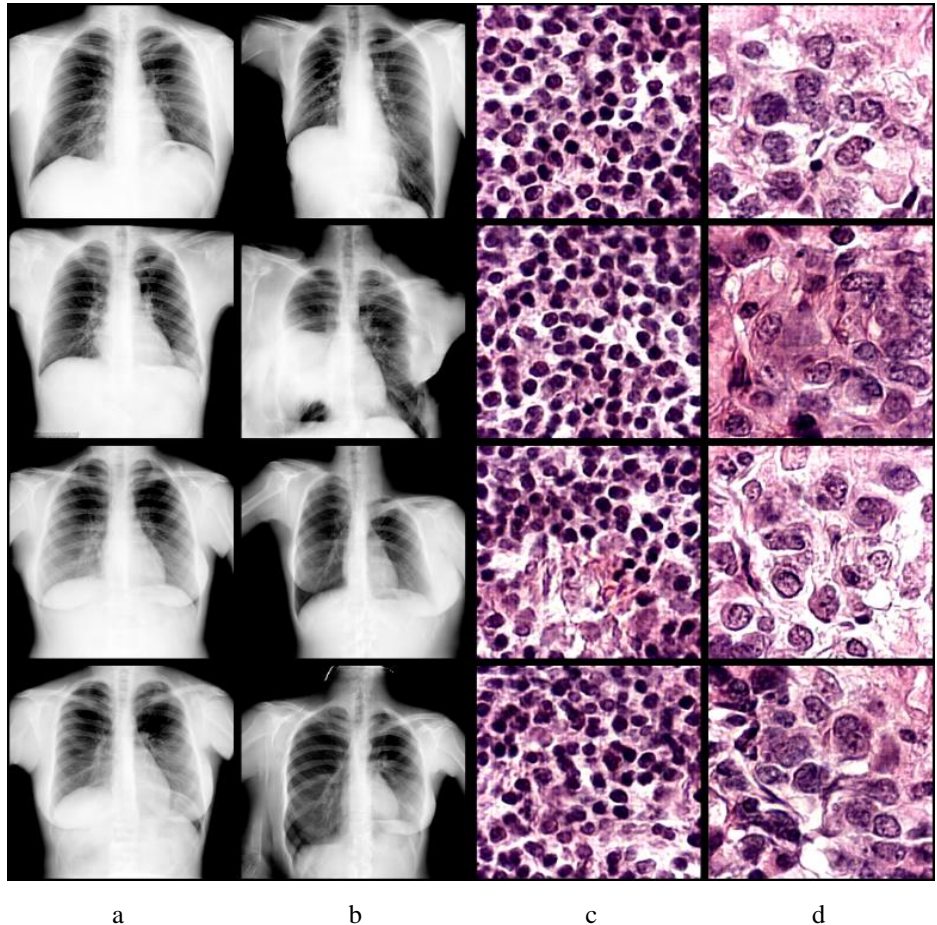

       a                     b                    c                    d

**Fig. 1.** Examples of artificial images created using DCGANS.
(a) Visually appealing chest X-ray images. (b) Heavily distorted X-ray images. (c) Histology images of normal tissue. (d) Histology images of malignant tumors.

However, in our view, the performance drop seems to be acceptable for many practical applications and there is still some potential for further improvements.

As it can be seen from the Table 3, both relative and the absolute accuracy drop for DL models is about 2 to 5 times lower than for conventional methods. Since conventional models work with LBP features while DL models treat image "as is", we can assume that generated images may be not very realistic in terms of LBP features. However, still they are much more similar when considered in the feature space of deep convolutional neural networks.

## 5      Conclusions

Results on generating artificial images of two different biomedical modalities obtained using two kinds of generative DL models allow drawing the following conclusions.

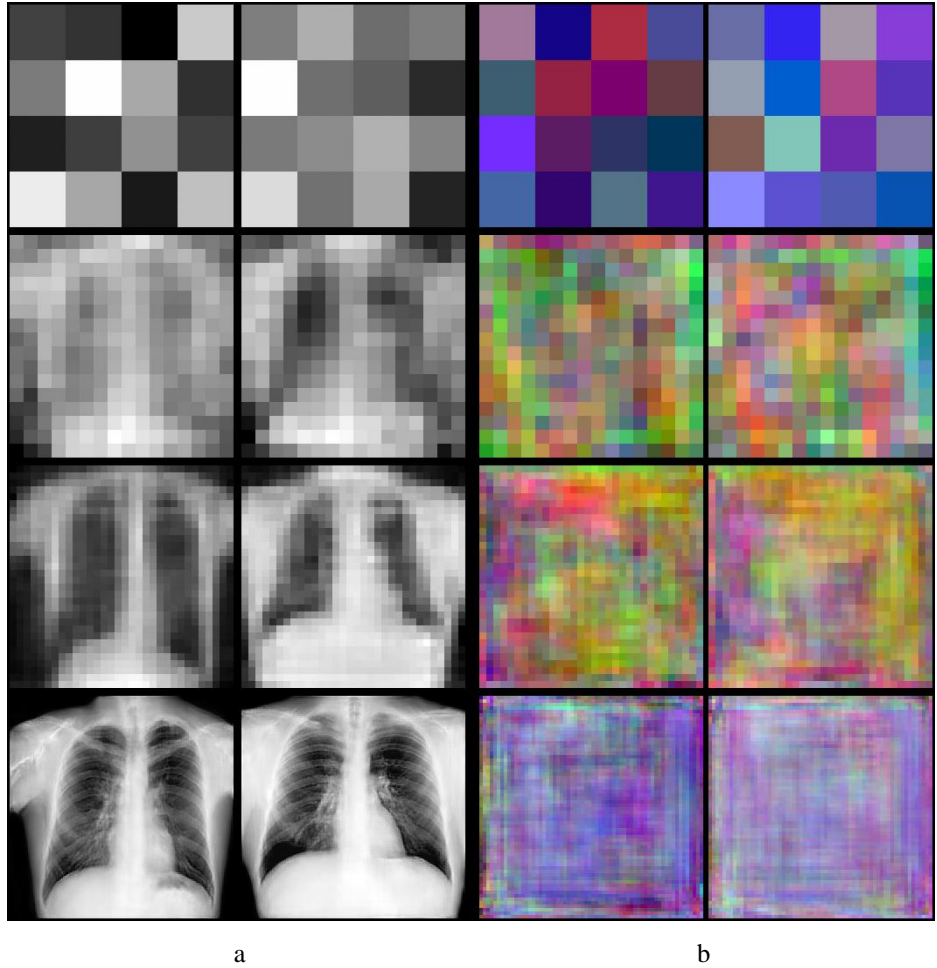

                           a                                                       b

**Fig. 2.** Examples of generation of artificial images by progressive refinement (from top to bottom) using PROGAN. (a) Chest X-ray images. (b) Histology images.

(1) Artificial X-ray and histology image datasets have a potential to be used as a substitute for real images in the training of classification models of different kinds. The relative accuracy drop was rather acceptable in the majority of benchmarks and ranged between 2.2% and 3.49% in case of DL and between 5.5% and 13.25% in case of conventional methods respectively.

(2) Artificial training data work substantially better when used in recent DL models comparing to the old-fashion models trained on classical LBP features. Probably, this can be generalized to other feature types and explained by conceptual similarity of image presentation in convolutional networks and natural difference between the latter and presentation by any set of conventional features.

(3) We could suggest that progressive GAN architectures may be recommended for generating anatomical images like the chest X-rays while more simple architectures such as DCGANs are better suited for color texture images like the histology ones.

**Table 3.** Benchmarking results (the best scores).

| Classification task | ML model | Accuracy, trained on real images | Accuracy, trained on artificial images | Relative accuracy drop |
|---|---|---|---|---|
| Histology: norm and tumor | VGG16 | **0.96** | **0.93** | **3.12%** |
| | K-NN | 0.94 | 0.87 | 7.45% |
| | SVM | 0.90 | 0.85 | 5.56% |
| | Random forest | 0.93 | 0.86 | 7.53% |
| X-ray: by subject's gender | VGG16 | **0.90** | **0.88** | **2.22%** |
| | K-NN | 0.84 | 0.78 | 7.14% |
| | SVM | 0.82 | 0.74 | 9.76% |
| | Random forest | 0.83 | 0.72 | 13.25% |
| X:ray: by subject's age group | VGG16 | **0.86** | **0.83** | **3.49%** |
| | K-NN | 0.80 | 0.73 | 8.75% |
| | SVM | 0.79 | 0.71 | 10.13% |
| | Random forest | 0.80 | 0.70 | 12.50% |